\documentclass[sigconf,nonacm]{acmart}
\renewcommand\footnotetextcopyrightpermission[1]{}

\usepackage{graphicx}
\usepackage{stfloats}
\usepackage{wrapfig}
\usepackage{lipsum}  
\usepackage[capitalise]{cleveref}

\newlength\savewidth

\usepackage{xcolor}

\usepackage{algorithm}
\usepackage[noend]{algpseudocode}

\algrenewcommand\alglinenumber[1]{#1}

\usepackage{soul}

\usepackage{multirow}
\usepackage[normalem]{ulem}
\useunder{\uline}{\ul}{}

\usepackage{enumitem}
\setlist[itemize]{leftmargin=*, topsep=.0em, itemsep=0pt, parsep=0pt, partopsep=0pt}
\setlist[enumerate]{leftmargin=*, topsep=.0em, itemsep=0pt, parsep=0pt, partopsep=0pt}

\usepackage[frozencache,cachedir=.]{minted}

\usepackage{amsthm}
\newtheoremstyle{def_style}
  {0.5em}     
  {0.5em}     
  {}          
  {}          
  {\bfseries} 
  {.}         
  {0.5em}      
  {}          

\theoremstyle{def_style}

\theoremstyle{def_style}

\usepackage{amsmath}
\usepackage{bbm}

\usepackage{collectbox}



\usepackage{mathtools}

\usepackage{balance}

\usepackage{caption}
\usepackage{subcaption}

\newcommand{\qnum}[1]{%
  \ifmmode
    \text{\textquotesingle}#1\text{\textquotesingle}%
  \else
    \textquotesingle#1\textquotesingle%
  \fi
}

\copyrightyear{2026}
\acmYear{2026}
\setcopyright{cc}
\setcctype{by}
\acmConference[KDD 2026] {Proceedings of the 32nd ACM SIGKDD Conference on Knowledge Discovery and Data Mining V.1}{August 9--13, 2025}{Jeju Island, Republic of Korea.}
\acmBooktitle{Proceedings of the 32nd ACM SIGKDD Conference on Knowledge Discovery and Data Mining V.1 (KDD 2026), August 9--13, 2025, Jeju Island, Republic of Korea}
\acmISBN{979-8-4007-2258-5/2026/08}
\acmDOI{10.1145/3770854.3783942}

\begin{document}

\newcommand{\proposed}{TREASURE}
\newcommand{\proposedfull}{\underline{TR}ansformer \underline{E}ngine \underline{A}s \underline{S}calable \underline{U}niversal transaction \underline{R}epresentation \underline{E}ncoder}

\title[The Visa Payment Foundation Model]{\proposed{}: The Visa Payment Foundation Model\\for High-Volume Transaction Understanding}

\author{Chin-Chia Michael Yeh}
\authornote{miyeh@visa.com}
\author{Uday Singh Saini}
\author{Xin Dai}
\author{Xiran Fan}
\author{Shubham Jain}
\affiliation{%
  \institution{Visa Research}
  \city{Foster City}
  \state{CA}
  \country{USA}
}

\author{Yujie Fan}
\author{Jiarui Sun}
\author{Junpeng Wang}
\author{Menghai Pan}
\author{Yingtong Dou}
\affiliation{%
  \institution{Visa Research}
  \city{Foster City}
  \state{CA}
  \country{USA}
}

\author{Yuzhong Chen}
\author{Vineeth Rakesh}
\author{Liang Wang}
\author{Yan Zheng}
\author{Mahashweta Das}
\affiliation{%
  \institution{Visa Research}
  \city{Foster City}
  \state{CA}
  \country{USA}
}

\renewcommand{\shortauthors}{Chin-Chia Michael Yeh et al.}

\begin{abstract}
\sloppy
Payment networks form the backbone of modern commerce, generating high volumes of transaction records from daily activities.
Properly modeling this data can enable applications such as abnormal behavior detection and consumer-level insights for hyper-personalized experiences, ultimately improving people's lives.
In this paper, we present \proposed{}, \proposedfull{}, a multipurpose transformer-based foundation model specifically designed for transaction data.
The model simultaneously captures both consumer behavior and payment network signals (such as response codes and system flags), providing comprehensive information necessary for applications like accurate recommendation systems and abnormal behavior detection.
Verified with industry-grade datasets, \proposed{} features three key capabilities:
1) an input module with dedicated sub-modules for static and dynamic attributes, enabling more efficient training and inference;
2) an efficient and effective training paradigm for predicting high-cardinality categorical attributes; and
3) demonstrated effectiveness as both a standalone model that increases abnormal behavior detection performance by 111\% over production systems and an embedding provider that enhances recommendation models by 104\%.
We present key insights from extensive ablation studies, benchmarks against production models, and case studies, highlighting valuable knowledge gained from developing \proposed{}.
\end{abstract}





\maketitle

\section{Introduction}
Visa operates as a global payment network company processing over 300B transactions or 15T dollars annually between more than 4B credentials and 150M+ merchants across 200+ countries~\cite{visa_factsheet}.
Visa deploys 100+ machine learning models that leverage payment network data to ensure transaction integrity and provide valuable services to all parties involved in these transactions~\cite{visa_intelligent}.
These machine learning models generally achieve their goals by modeling both cardholder behavior and payment network signals for each transaction.
For example, the Stand-in-Processing (STIP) service determines authorization decisions when issuing banks are unavailable to do so~\cite{visa_stip}.
The machine learning model utilized for STIP service needs to distinguish between normal and abnormal cardholder behaviors and predict what signals (i.e., approved or declined) the payment network would send to merchants.

In this paper, we explore the design of a unified foundation model that addresses multiple transaction-related machine learning tasks across multiple sectors.
We call our proposed model \proposed{}, short for \proposedfull{}.
The \proposed{} model effectively captures both cardholder behaviors and various payment network signals at the scale of Visa's data.
To achieve this goal, \proposed{} incorporates three key capabilities:
1) an input module with dedicated sub-modules for static and dynamic attributes, enabling more efficient training and inference;
2) an efficient and effective training paradigm for predicting high-cardinality categorical attributes; and
3) demonstrated effectiveness both as a standalone system for transaction-based tasks and as an embedding provider for downstream applications.

\begin{figure}[htp]
\centerline{
\includegraphics[width=0.99\linewidth]{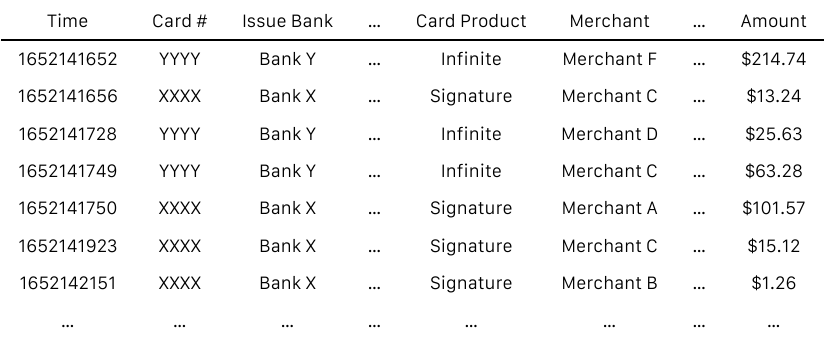}
}
\caption{
Example of raw transaction data showing interleaved transactions from different cards.
}
\label{fig:example}
\end{figure}

The first capability addresses the inherent characteristics of transaction data.
As shown in \cref{fig:example}, raw transaction data is logged as individual entries whenever transactions occur, with all attributes potentially changing between transactions.

\begin{figure}[htp]
\centerline{
\includegraphics[width=0.99\linewidth]{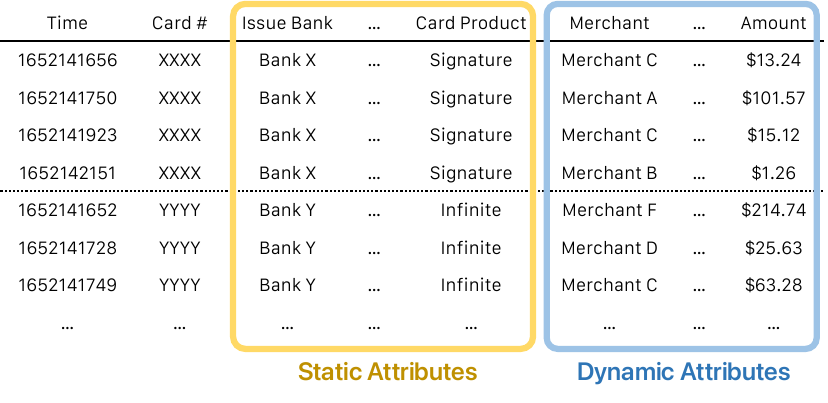}
}
\caption{
Grouped transactions from the same card, demonstrating that attributes can be either static or dynamic when considering the shopping behavior of each cardholder.
}
\label{fig:example_group}
\end{figure}

However, since our goal is to model cardholder behavior, it is more meaningful to group transactions from the same card together and order them chronologically.
After grouping and sorting, as illustrated in \cref{fig:example_group}, we observe that certain attributes remain constant throughout a card's transaction history, while others vary with each transaction.
We refer to the unchanging card-associated attributes as \textit{static} attributes, and the changing ones as \textit{dynamic} attributes.
As demonstrated in \cite{zhang2023fata}, modeling different types of attributes using specialized sub-modules proves both more effective and efficient.
Therefore, this first capability is crucial in our development of a foundation model for transaction data.

The second capability addresses the challenges in training the foundation model.
As the model learns cardholder behavior through next-transaction prediction tasks, it must predict all attributes associated with a transaction.
This includes categorical attributes with extremely high cardinality.
For example, predicting the next merchant using cross-entropy loss would require computing over 150M+ logits for each prediction.
Compared to large language models with vocabulary sizes typically under one million~\cite{hugging_llama4,hugging_deepseek,ji2024gemma}, our problem is significantly more computationally expensive.
Consequently, we cannot simply apply standard cross-entropy loss and must develop a more efficient and effective training paradigm for predicting high-cardinality categorical attributes.
In our paper, we provide an efficient solution for approximating cross-entropy loss in the design of \proposed{}.

The third capability demonstrates \proposed{}'s effectiveness in real-world applications, validating its practical utility.
We have tested \proposed{} in two distinct scenarios: 1) as a standalone model and 2) as an embedding service provider.
When used as a standalone model, we evaluate its abnormal behavior detection capabilities against production models currently deployed in the field, measuring improvements in detection accuracy.
Our experiments show that \proposed{} can identify suspicious patterns that traditional models might miss, particularly increasing performance by 111\%.
When used as an embedding provider, we demonstrate how embeddings generated from transaction history can enhance recommendation system performance by 90\% through capturing nuanced financial behavior patterns.
These embeddings encapsulate temporal spending habits, merchant category preferences, and transaction value distributions, providing a rich foundation for downstream applications without requiring access to the raw transaction data.
This dual-purpose capability makes \proposed{} particularly valuable as it can serve both specialized security functions and broader consumer experience applications with the same underlying model architecture.

Our key contributions are as follows:
\begin{itemize}
    \item Identification of key capabilities required for a transaction foundation model and development of corresponding solutions.
    \item Analysis of various design choices for both model and dataset construction, including the scaling law for transaction data.
    \item Demonstration of \proposed{}'s superior performance both as a standalone model and as an embedding service.
\end{itemize}
\section{Related Work}
Foundation models, defined as models trained at scale on vast datasets, have become a cornerstone of modern AI research~\cite{bommasani2021opportunities}.
While large language models (LLMs) like GPT are prominent examples~\cite{brown2020language}, the foundation model paradigm has been successfully adapted to other domains, including computer vision~\cite{alayrac2022flamingo,radford2021learning}, time series~\cite{yeh2023toward,das2024decoder,rasul2023lag}, and tabular data~\cite{van2024tabular,yang2023unitabe}.
Our work, \proposed{}, builds on these advancements, situating itself at the intersection of foundation models for tabular and sequential transaction data~\cite{skalski2023towards}.

General-purpose tabular foundation models, such as those in~\cite{van2024tabular,yang2023unitabe}, are designed to learn relationships between columns within a static table.
However, they are not inherently equipped to model the critical sequential dependencies between rows that characterize transaction data, as illustrated in \cref{fig:example_group}.
The temporal patterns in transaction sequences reflect consumer behavior, which is vital for key applications in the payment industry.
Consequently, while these models offer broad applicability, their lack of a sequential focus makes them suboptimal for our specific use case.
We do, however, concur with the analysis in~\cite{van2024tabular} regarding the limitations of directly applying LLMs to tabular data, namely their inefficiency in modeling continuous variables and their significant computational expense.

The work most closely related to ours is the transaction foundation model proposed in~\cite{skalski2023towards}, which also treats transactions as a sequence of events.
Despite this similarity, a key distinction sets our work apart: their model does not incorporate signals from the payment network itself.
This design choice means their model cannot leverage direct indicators from the payment network, such as issuer decline codes, which provide powerful, real-time evidence that significantly enhances tasks like abnormal behavior detection.
Furthermore, their approach relies on Gated Recurrent Units (GRUs)~\cite{cho2014learning}, an architecture that our experiments in \cref{sec:temporal_exp} demonstrate to be inferior to the Transformer~\cite{vaswani2017attention} for modeling transaction sequences.
To the best of our knowledge, \proposed{} is the first Transformer-based foundation model designed to holistically model payment transactions by simultaneously capturing both consumer behavior and payment network signals, validated at an industrial scale.
\section{Methodology}
This section details the proposed \proposed{} model.
We begin by outlining the input and output data structures in \cref{sec:intput_output}.
Subsequently, we present the overall model architecture in \cref{sec:overall}.
The core challenge in developing a foundation model for transaction data is the effective modeling of diverse attributes; thus, we dedicate specific subsections to the input module (\cref{sec:input_module}) and the output module (\cref{sec:output_module}).
Finally, we describe the optimization objectives in \cref{sec:optimization}.

\subsection{Input and Output}
\label{sec:intput_output}
Each data sample comprises an input sequence and two corresponding output sequences, all derived from the transactions associated with a single card.
For a card with $t$ transactions, the input sequence consists of a static attribute vector ($X_s$) and $t$ dynamic attribute vectors ($X_{d,1}, \dots, X_{d,t}$), where each $X_{d,i}$ corresponds to the $i$-th transaction.
Raw attributes are classified as either static or dynamic.
Each attribute vector (static or dynamic) contains both numerical and categorical attributes.
Numerical attributes are stored as floating-point numbers representing their raw values.
Categorical attributes are stored as integers, representing category indices derived from predefined mappings.

For instance, if a transaction amount is \$3.14, the value 3.14 is stored as a floating-point number in the corresponding $X_{d,i}$.
If the same transaction includes the numeric-3 country code \texttt{840} (United States of America, per ISO 3166-1 numeric~\cite{wikipedia_iso3166_1_numeric}), an integer value, such as 58, obtained from a mapping table (e.g., $\{\dots, \qnum{834}:57, \qnum{840}:58, \qnum{850}:59, \dots\}$), is stored in $X_{d,i}$.
Given 249 unique numeric-3 country codes, this specific mapping table would map each code to an integer in the range $[0, 248]$, where the range size equals the attribute's cardinality.

For each of the $t$ transactions, \proposed{} generates two sets of outputs.
The first output set predicts the attributes of the \textit{next} transaction (i.e., transaction $i+1$).
The second output set predicts payment network signals associated with the \textit{current} transaction (i.e., transaction $i$).
These network signals are used exclusively as outputs, as they represent outcomes of a transaction (e.g., a response code indicating approval or decline) that are only known \textit{after} it has been processed. 
Therefore, given the input at time step $i$, the model performs two tasks: the \textit{current} prediction head infers the signals for the transaction at time step $i$, while the \textit{next} prediction head forecasts the attributes for the transaction at time step $i+1$.

\subsection{Overall Architecture}
\label{sec:overall}
\begin{figure}[htp]
\centerline{
\includegraphics[width=0.85\linewidth]{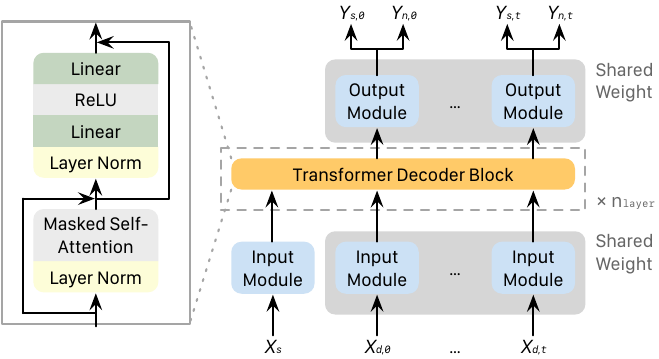}
}
\caption{
The overall model architecture of \proposed{}.
}
\label{fig:overall}
\end{figure}

With the input and output structures outlined, we now detail the overall architecture of \proposed{}, illustrated in \cref{fig:overall}.
The model processes two input types—static and dynamic—using two distinct input modules of identical design.
The static input module processes the single static vector $X_s$.
The dynamic input module processes each dynamic vector $X_{d,i}$ in the sequence, with its weights shared across all time steps (transactions).
The design of these input modules is further detailed in \cref{sec:input_module}.

After processing by their respective input modules, the resulting intermediate representations are fed into a Transformer decoder block.
This block employs causal masked self-attention to capture temporal dependencies within the transaction sequence.
To ensure that dynamic transaction vectors can attend to the static information, the static vector's representation is positioned first in the sequence supplied to the Transformer, facilitated by the causal mask.
The Transformer decoder block's design is depicted in the inset of \cref{fig:overall}.
We omit explicit positional encoding, as a decoder-only Transformer architecture inherently captures the relative ordering of transactions through its autoregressive nature and causal masking~\cite{kazemnejad2023impact,irie2024positional}.

For each dynamic vector's representation processed by the Transformer, an output module generates the final predictions.
Similar to the dynamic input module, the output module's weights are shared across time steps.
This output module contains two prediction heads, each dedicated to one of the two output types (the next transaction attributes~$Y_{n,i}$ and the current transaction signals~$Y_{s,i}$, where $i \in \{0, \dots, t\}$).
The detailed design of the output module is presented in \cref{sec:output_module}.

\subsection{Input Module}
\label{sec:input_module}
\begin{figure}[htp]
\centerline{
\includegraphics[width=0.6\linewidth]{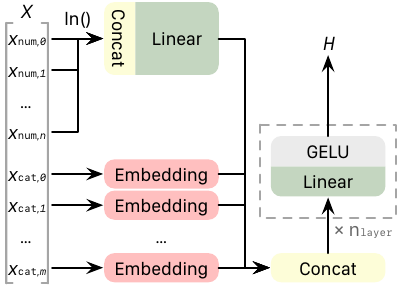}
}
\caption{
The detailed input module of \proposed{}.
$H$ represents the input to the Transformer decoder block.
}
\label{fig:input}
\end{figure}

The detailed architecture of the input module is illustrated in \cref{fig:input}.
Its primary function is to model interdependencies among different input attributes within a single transaction vector (either static or one dynamic vector).

Numerical and categorical attributes are processed differently.
Numerical attributes are first transformed to a logarithmic scale, as all numerical features in our dataset (e.g., transaction amounts, time differences between transactions) exhibit long-tail distributions.
These log-scaled numerical attributes are then concatenated into a single numerical input vector, which is subsequently processed by a linear layer.

For categorical attributes, each raw value is initially converted to a category index, as described in \cref{sec:intput_output}.
This index is directly used to retrieve a corresponding embedding from an attribute-specific embedding table.\footnote{In subsequent work, we demonstrated that initializing \proposed{}'s categorical embeddings with LLM-based sentence embeddings can further improve model performance~\cite{fan2025enhancing}.}
We have implemented mechanisms to handle new category values for categorical attributes that may appear during inference but were not present during training.
Such OOV (out-of-vocabulary) values can arise from new merchants, emerging business categories, or data quality variations.
These mechanisms ensure that an embedding index is consistently available for every categorical attribute value.
However, the specific implementation details are omitted to protect sensitive information about the raw category values.

Finally, the representation vector from the numerical linear layer and the embedding vectors for all categorical attributes are concatenated.
This combined vector is then processed through additional linear and activation layers to produce a unified representation that captures information across all input attributes.

\subsection{Output Module}
\label{sec:output_module}
\begin{figure}[htp]
\centerline{
\includegraphics[width=0.65\linewidth]{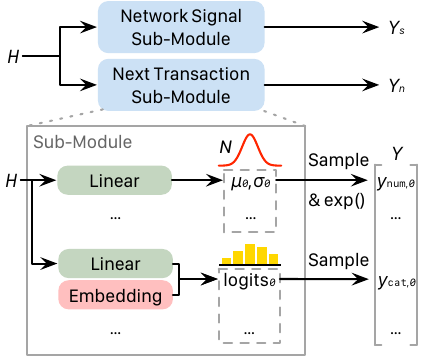}
}
\caption{
The detailed output module of \proposed{}. 
Two identical sub-modules predict current signals and next transaction attributes using the output of Transformer decoder block, $H$, as their input.
}
\label{fig:output}
\end{figure}

The detailed structure of the output module is depicted in \cref{fig:output}.
This module generates predictions corresponding to each transaction representation output by the Transformer block.
It comprises two sub-modules, one for predicting \textit{next} transaction attributes and one for predicting \textit{current} transaction network signals.
These sub-modules share an identical architecture, as both must predict numerical and categorical values.
They differ only in the specific attributes they target and the temporal origin of their ground truth data.

The input to the output module is the hidden representation $H$ from the Transformer block for a given time step. 
For numerical attributes, the module predicts a full probability distribution rather than a single point estimate, allowing it to capture the inherent uncertainty in its predictions. 
This is achieved by estimating the mean ($\mu$) and standard deviation ($\sigma$) of the target value in logarithmic scale. 
This approach effectively models numerical attributes using log-normal distributions---a choice motivated by their long-tail nature in our data~\cite{shchur2019intensity}. 
A final predicted value can be obtained by sampling from this estimated distribution $N(\mu, \sigma)$ and then exponentiating the result.

For categorical attributes, the representation $H$ is first transformed by an attribute-specific linear layer.
Thus, if predicting 10 distinct categorical attributes, 10 such linear layers are employed.
The output of this linear layer is then used to compute logits for each possible category of that attribute via a vector-matrix multiplication with the attribute's corresponding embedding table.
For instance, to predict one of 249 numeric-3 country codes, the logits are computed by multiplying the output of the country-code-specific linear layer with the country code embedding matrix.
A predicted categorical value can be obtained by sampling based on these computed logits.

\subsection{Optimization Objective}
\label{sec:optimization}
The learning objective of \proposed{} is to accurately predict numerical and categorical attributes for both next-transaction attributes and current-transaction network signals.
Consequently, the overall loss function is an aggregation of individual loss terms, each corresponding to a specific predicted attribute.
We first define the loss calculation for different attribute types before presenting the aggregate loss.

Attributes are categorized into three types for loss computation: 1) numerical attributes, 2) low-cardinality categorical attributes, and 3) high-cardinality categorical attributes.
The distinction between low and high cardinality is determined by a predefined threshold, which is set to 1,024 in this work.
In our dataset, categorical attributes range in cardinality from 2 to over 100M.
For high-cardinality attributes prone to long-tail distributions and out-of-vocabulary items, we employ a principled aggregation strategy where infrequent or new entities are mapped to shared aggregated identifiers, controlling vocabulary size while maintaining attribute utility.
For numerical attributes, we employ the negative log-likelihood of a normal distribution, where the target variable is in logarithmic scale:
\begin{equation} \scriptsize
    \mathcal{L}_\text{num}(\mu, \sigma, y) = \frac{(y - \mu)^2} {2 \sigma^2} + \log(\sigma) + \frac{\log(2\pi)}{2}
\end{equation}
\noindent Here, $\mu$ (mean) and $\sigma$ (standard deviation) are model outputs, and $y$ is the ground truth value, all in logarithmic scale.

For low-cardinality categorical attributes, we use the standard cross-entropy loss:
\begin{equation}
\label{eq:lcat} \scriptsize
    \mathcal{L}_\text{lcat}(Z, y) = -\log{\frac{e ^ {Z[y]}}{\sum^C_{i=1}e^{Z[i]}}}
\end{equation}
\noindent where $Z \in \mathbb{R}^C$ is the vector of logits for $C$ categories, and $y$ is the index of the ground truth category.

For high-cardinality categorical attributes, direct computation of \cref{eq:lcat} is infeasible due to the excessive GPU memory required to compute logits for all categories for every attribute, time step, and sample.
Instead, we use the InfoNCE loss~\cite{oord2018representation}, as defined in \cref{eq:hcat}.
This approach requires computing the logit for only the positive (ground truth) category and a subset of negative categories:
\begin{equation}
\label{eq:hcat} \scriptsize
\mathcal{L}_{\text{hcat}}(H_\text{a}, \mathbf{E}, y, I) = -\log \frac{e^{H_\text{a} \cdot \mathbf{E}[y, :]}}{\sum_{i\in I} e^{H_\text{a} \cdot \mathbf{E}[i, :]}} 
\end{equation}
\noindent where $H_\text{a}$ is the output of the attribute-specific linear layer in the output module, $\mathbf{E}$ is the embedding matrix for the categorical attribute, $y$ is the positive category's index, and $I$ is a set containing the index $y$ and indices of sampled negative categories.

Sharing negative indices across all time steps and samples within a batch significantly reduces memory consumption. The pseudocode in \cref{alg:hcat} provides a concrete illustration of this high-cardinality loss computation and its sharing mechanism.

\begin{algorithm}[htp]
\caption{The function computes loss for high-cardinality categorical attributes.}\label{alg:hcat}
\begin{minted}[breaklines,xleftmargin=2em,linenos,fontsize=\scriptsize, escapeinside=!!]{python}
def high_cardinality_loss(hidden, embedding, label_positive, n_negative):
    """
    Input:
        hidden: (batch_size, sequence_length, hidden_dimension)
        embedding: (n_category, hidden_dimension)
        label_positive: (batch_size, sequence_length)
        n_negative: int
    Output:
        loss: (batch_size, sequence_length)
    """
    hidden_positive = embedding[label_positive, :]
    label_negative = torch.randint(
        embedding.shape[0], (n_negative, ),
        dtype=torch.long, device=label_positive.device)
    hidden_negative = embedding[label_negative, :]
    dot_positive = torch.einsum(
        'ijk,ljk->ilj', hidden, hidden_positive)
    dot_negative = torch.einsum(
        'ijk,lk->ilj', hidden, hidden_negative)
    dot_all = torch.cat([dot_positive, dot_negative], dim=1)
    indices = torch.arange(hidden.shape[0])
    loss_numerator = dot_positive[indices, indices]
    loss_denominator = torch.logsumexp(dot_all, dim=1)
    return loss_numerator - loss_denominator
\end{minted}
\end{algorithm}

\begin{figure}[htp]
\centerline{
\includegraphics[width=0.99\linewidth]{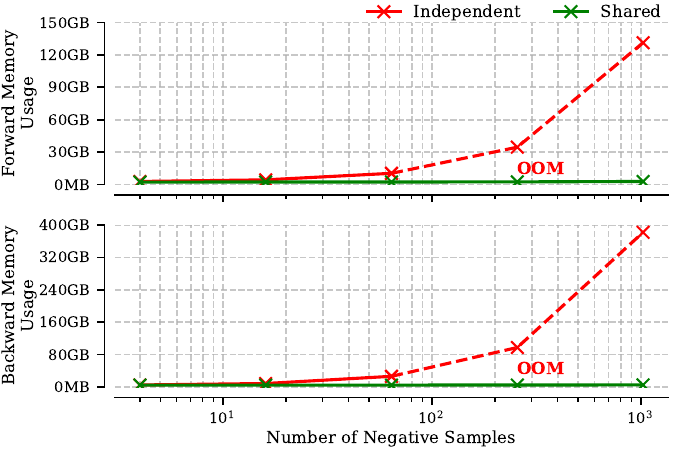}
}
\caption{
Efficiency improvement through shared negative sampling in \proposed{}.
By sharing negative samples across both samples and time steps within a batch, we dramatically reduce the memory usage during training.
Dashed lines indicate linear extrapolations from the solid lines.
}
\label{fig:hcat_exp}
\end{figure}

To validate the efficacy of our shared negative sampling strategy, we compared its memory footprint against a baseline using independent negative samples for each loss computation.
Memory usage for both forward and backward passes was measured across varying numbers of negative samples, as shown in \cref{fig:hcat_exp}.
The experiment was conducted with PyTorch 2.6.
The shared sampling method maintained forward pass memory below 3GB and backward pass memory below 6GB.
In contrast, the independent sampling method encountered out-of-memory errors when the number of negative samples exceeded 64.
The memory usage for configurations resulting in out-of-memory errors is extrapolated linearly from the last successful runs.

With individual attribute loss terms defined, we now describe their aggregation into the overall loss.
The abnormal behavior flag is considered the most critical attribute, indicating transaction normality, i.e., whether a transaction should be considered part of a user's normal behavior.
Its corresponding loss, $\mathcal{L}_\text{abnormal}$, serves as a reference to scale other attribute losses.
The overall loss $\mathcal{L}$ is formulated as:
\begin{equation}
\label{eq:overall} \scriptsize
\mathcal{L} = \mathcal{L}_\text{abnormal} + \frac{1}{|\mathbf{L}|} \sum_{\mathcal{L}_i \in \mathbf{L}}{\min\left(\mathcal{L}_i, \frac{\mathcal{L}_i \hat{\mathcal{L}}_\text{abnormal}}{\hat{\mathcal{L}}_i}\right)}
\end{equation}
\noindent where $\mathbf{L}$ is the set of all non-abnormal behavior attribute losses.
Loss terms with a hat (e.g., $\hat{\mathcal{L}}_\text{abnormal}, \hat{\mathcal{L}}_i$) denote detached values, meaning gradients are not computed through them when they are used in this scaling mechanism.

The first term is the direct loss for the abnormal behavior flag.
The second term aggregates the other losses, dynamically adjusting their contributions.
Specifically, for each auxiliary loss $\mathcal{L}_i$, its effective value in the sum is $\mathcal{L}_i$ if $\hat{\mathcal{L}}_i$ is smaller than or equal to $\hat{\mathcal{L}}_\text{abnormal}$ (i.e., task $i$ is performing better or similarly).
If task $i$ is performing worse (i.e., $\hat{\mathcal{L}}_i > \hat{\mathcal{L}}_\text{abnormal}$), its current loss $\mathcal{L}_i$ is scaled down by the ratio $\hat{\mathcal{L}}_\text{abnormal}/\hat{\mathcal{L}}_i$.
This dynamic loss scaling mechanism ensures that the primary abnormal behavior detection task dominates the overall gradient direction while the auxiliary tasks continue to contribute meaningfully to the learning process. 
By prioritizing the abnormal behavior flag, the model focuses on the most critical attribute while still leveraging auxiliary tasks to improve generalization and representation learning.

\section{Experiment}
In this section, we first present our experiment setup in \cref{sec:dataset}.
Next, in \cref{sec:optimization_exp}, we validate the benefits of a multipurpose foundation model compared to single-purpose models and the efficacy of our adopted loss aggregation strategy.
We then compare the Transformer backbone used in \proposed{} with RNN-based backbones commonly employed in transaction models in \cref{sec:temporal_exp}.
Following this, in a dedicated subsection, we evaluate the effectiveness of different negative sampling strategies.
Subsequently, we demonstrate the capability of \proposed{} to serve embeddings for specialized applications through a case study in \cref{sec:serving}.
We also present visualizations of the learned embeddings in \cref{sec:visual}.
Finally, in \cref{sec:scaling_exp}, we demonstrate that the performance of \proposed{} improves with scale, both in terms of training data volume and model size.

\subsection{Dataset and Experiment Setup}
\label{sec:dataset}
We sampled approximately six billion transactions from 30 million distinct cardholders, recorded between September 1, 2020, and November 30, 2022.
This dataset spans a total of 26 months.
Transactions from the initial 24 months constitute the training data, the 25th month serves as the validation data, and the final month is used as the test data.
In the sampled dataset, each transaction comprises five static attributes and sixteen dynamic attributes.
During training, \proposed{} predicts all sixteen dynamic attributes of the subsequent transaction and two network signals associated with the current transaction.
The specific names of these attributes cannot be disclosed due to the sensitive nature of the project.
\proposed{} uses a 3-layer Transformer decoder with 4 attention heads and a hidden dimension of 256.
The input module also uses 3 layers.
The model is trained for 20 epochs using the AdamW optimizer with a learning rate of $10^{-4}$, default beta parameters, and a batch size of 256.
The best checkpoint is selected based on validation performance.

We assess the model's performance in predicting the next transaction using precision at one (Prec@1) for selected categorical attributes (i.e., merchant, its country, city, and category) and symmetric mean absolute percentage error (sMAPE) for a selected numerical attribute (i.e., transaction amount).
While we report results for selected attributes with high business value and interpretability, our complete evaluations across all attributes show consistent trends.
Additionally, we measure the model's performance in abnormal behavior detection using an in-house performance metric.
Detailed performance figures cannot be disclosed due to the sensitive nature of this task.
To present these results, we compute the performance ratio between the evaluated model and the currently deployed system.
For instance, a ratio of 1.5 signifies that the evaluated model is 50\% better than the currently deployed system in abnormal behavior detection.
We term this metric Relative Improvement (RI).

For deployment considerations, we cap the sequence length at 512 transactions, which covers over two years of transaction history for most cardholders and provides a deterministic computational budget.
Under this configuration, the model's inference latency remains operationally feasible for real-time transaction processing.
The deployment of machine learning models in critical financial infrastructure follows a rigorous multi-stage process involving extensive offline evaluation, parallel validation phases, and regulatory review.
The comprehensive offline evaluation results presented in this paper represent a significant milestone in this process, and we are actively working toward full production deployment.

\subsection{Optimization Objective}
\label{sec:optimization_exp}
In the first set of experiments, we pursued two primary objectives:
1) to validate the utility of a multipurpose foundation model for transaction data over specialized single-purpose models, and
2) to confirm the necessity of the adopted loss aggregation strategy, i.e., \cref{eq:overall}.
To achieve the first objective, we trained two single-purpose baseline models.
The first baseline focuses on abnormal behavior detection, while the second concentrates on predicting the most likely next merchant.
A model's ability to predict the next merchant serves as a proxy for its performance in recommendation applications.
To achieve the second objective, we trained two additional models employing alternative loss aggregation strategies: 1) simple summation of losses, and 2) weighting losses for equal contribution.
The experiment results are presented in \cref{tab:optimization}.

\begin{table}[htp]
\caption{Performance comparisons across different optimization objectives.
Arrows indicate improvement direction.
Bold and underlined values represent best and second-best results, respectively.}
\label{tab:optimization}
\begin{center}
\resizebox{0.99\linewidth}{!}{%
\begin{tabular}{ll||cc|cc|c}
\multirow{2}{*}{Attribute} & \multirow{2}{*}{Measure} & \multicolumn{2}{c|}{Single Purpose} & \multicolumn{3}{c}{Multiple Purposes}            \\ \cline{3-7}
                           &                          & Merchant       & Abnormal          & Simple       & Equal           & \textbf{\proposed{}}        \\ \hline \hline
Abnormal                   & RI ($\uparrow$)                      & -              & {\ul 1.9606}      & 1.8768       & 0.1768          & \textbf{2.1171} \\ 
Amount                     & sMAPE ($\downarrow$)                    & -              & -                 & {\ul 0.5850} & 0.5893          & \textbf{0.5786} \\  \hline
Merchant                   & \multirow{4}{*}{Prec@1 ($\uparrow$)}                   & 0.1248         & -                 & {\ul 0.1306} & 0.1235          & \textbf{0.1421} \\
Country                    &                    & -              & -                 & {\ul 0.5587} & 0.5132          & \textbf{0.5634} \\
City                       &                    & -              & -                 & 0.1840       & \textbf{0.2050} & {\ul 0.1892}    \\
Category                   &                    & -              & -                 & {\ul 0.4291} & 0.3672          & \textbf{0.4335} \\
\end{tabular}%
}
\end{center}
\end{table}

Comparing the performance of \proposed{} with the two single-purpose models, we observe that our method simultaneously outperforms both.
This indicates that the proposed foundation model not only conserves resources by enabling a single multi-purpose model to replace multiple single-purpose ones but also delivers superior prediction quality.
These results demonstrate the advantages of \proposed{} compared to task-specific models.

When contrasting the loss aggregation strategy used in \proposed{} with the simple aggregation strategy, our adopted approach yields superior performance across all evaluated aspects.
This highlights the necessity of rebalancing the contributions of different loss terms.
Compared to the equal contribution strategy, the model trained with this alternative strategy shows improved performance in predicting the next merchant's city.
However, its performance on other attributes is inferior, even to that of the simple aggregation strategy.
Notably, the abnormal behavior detection capability, a critical application in payment systems, is significantly degraded by the equal contribution strategy.
One plausible explanation is that pivoting each loss term relative to the abnormal detection loss enhances the model's sensitivity to anomalies, thereby improving its ability to capture sequential patterns in transactions.

In conclusion, the multipurpose foundation model \proposed{} is more efficient and effective than multiple single-purpose models.
Furthermore, the adopted loss aggregation strategy is substantially more effective than the alternative baseline strategies.
Overall, \proposed{} demonstrates a significant RI, outperforming the currently deployed system by 111\%.

\subsection{Temporal Modeling Architecture}
\label{sec:temporal_exp}
While \proposed{} utilizes the Transformer architecture~\cite{vaswani2017attention}, existing foundation models for transaction data often employ GRUs~\cite{cho2014learning,skalski2023towards}.
In this set of experiments, we compare variants of \proposed{} that utilize different architectures for capturing temporal dependencies to validate the choice of Transformer modules in our method.
Specifically, we compared our Transformer-based solution with RNN-based alternatives (i.e., GRU~\cite{cho2014learning} and LSTM~\cite{hochreiter1997long}).
The experiment results are detailed in \cref{tab:temporal}.

\begin{table}[htp]
\caption{Performance comparisons across different temporal modeling architectures.
Arrows indicate improvement direction.
Bold and underlined values represent best and second-best results, respectively.}
\label{tab:temporal}
\begin{center}
\resizebox{0.99\linewidth}{!}{%
\begin{tabular}{l||cc|cccc}
            & Abnormal        & Amount          & Merchant        & Country         & City            & Category        \\ 
              & RI ($\uparrow$)              & sMAPE ($\downarrow$)           & \multicolumn{4}{c}{Prec@1 ($\uparrow$)}                                            \\ \hline \hline
\proposed{} + LSTM & {\ul 1.4427}    & {\ul 0.5850}    & 0.1103          & {\ul 0.5336}    & 0.1546          & {\ul 0.4003}    \\
\proposed{} + GRU  & 1.3979          & 0.6013          & {\ul 0.1172}    & 0.5291          & {\ul 0.1605}    & 0.3945          \\ \hline
\textbf{\proposed{}}        & \textbf{2.1171} & \textbf{0.5786} & \textbf{0.1421} & \textbf{0.5634} & \textbf{0.1892} & \textbf{0.4335}
\end{tabular}%
}
\end{center}
\end{table}

Both RNN-based solutions yield similar performance across all evaluation measures.
However, our Transformer-based solution consistently outperforms these RNN-based variants across all performance measures.
Therefore, the Transformer architecture proves to be a more effective choice for modeling temporal dependencies between transactions within the \proposed{} foundation model.

\subsection{Negative Sampling Strategy}
\sloppy
Efficient negative sampling is crucial for training models on large-scale transaction data, particularly for tasks involving high-cardinality categorical outputs.
In this subsection, we compare the effectiveness of different negative sampling strategies.
Given that \cref{sec:optimization_exp} demonstrated the significant computational burden associated with an independent sampling strategy when employing a large number of negative samples per positive instance, we use a smaller negative sampling size for this strategy in our current comparison.
For the shared negative sampling strategy employed in \proposed{}, we set the number of negative samples to 1024 per batch.
For the independent sampling strategy, we used five negative samples per positive instance.
The experiment results are presented in \cref{tab:negative}.

\begin{table}[htp]
\caption{Performance comparisons across different negative sampling strategies. 
The reported metric is Prec@1. 
Bold values indicate the best results.}
\label{tab:negative}
\begin{center}
\resizebox{0.7\linewidth}{!}{%
\begin{tabular}{l||cc|cc}
            & \multicolumn{2}{c|}{Low Cardinality} & \multicolumn{2}{c}{High Cardinality} \\ 
            & Country          & Category         & City              & Merchant         \\ \hline \hline
Independent & 0.5607           & 0.4324           & 0.1370            & 0.0600           \\
Shared      & \textbf{0.5634}  & \textbf{0.4335}  & \textbf{0.1892}   & \textbf{0.1421} 
\end{tabular}%
}
\end{center}
\end{table}

For low-cardinality attributes, the performance is comparable across different sampling methods.
This is anticipated, as the choice of negative sampling strategy has a negligible impact on the loss computation for attributes with few distinct values.
However, for high-cardinality attributes, we observe that the model trained with independent sampling exhibits significantly poorer performance.
This is likely attributable to the insufficient number of negative samples available during the loss computation for these attributes under the independent sampling regime.
Consequently, \proposed{} is trained using a large pool of negative samples generated on a per-batch basis (shared negative sampling), rather than performing sampling independently for each time step of every instance in a batch.

\subsection{Embedding Serving}
\label{sec:serving}
As \proposed{} is a foundation model, a key aspect of its versatility is its ability to provide high-quality embeddings for downstream tasks.
In this case study, we evaluate the utility of embeddings generated by \proposed{} in a merchant recommendation system.
Specifically, the task is to recommend the next merchant a user is likely to interact with.
We address this using a standard two-tower recommendation architecture~\cite{hu2008collaborative,koren2009matrix,wang2019neural,he2020lightgcn}, where one tower generates merchant embeddings and the other generates user embeddings.

\begin{figure}[htp]
\centerline{
\includegraphics[width=0.99\linewidth]{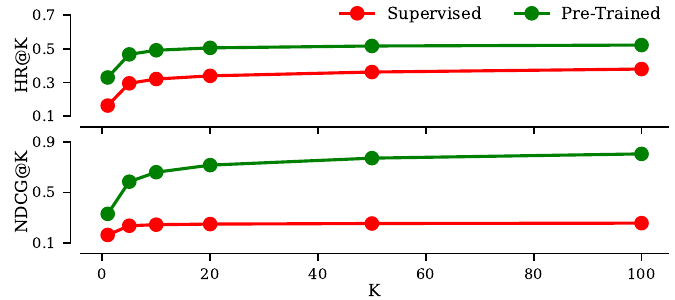}
}
\caption{
Embeddings generated by \proposed{} demonstrate superior effectiveness compared to embeddings from supervised training for recommendation tasks.
}
\label{fig:rec_exp}
\end{figure}

In these experiments, we compare user and merchant embeddings derived from \proposed{} against embeddings learned via supervised training tailored specifically for this recommendation task.
The training, validation, and test datasets for this case study were sampled from transactions occurring after September 1, 2022.
This ensures that \proposed{} had not encountered these specific user-merchant interactions during its pre-training phase, allowing for a fair evaluation of its generalization capabilities.
For clarity, this ``pre-training'' refers to the general-purpose foundation model training detailed in \cref{sec:optimization} and is distinct from the training performed specifically for the recommendation task.
Performance is measured using Hit Rate at $K$ (HR@$K$) and Normalized Discounted Cumulative Gain at $K$ (NDCG@$K$).
The experiment results are illustrated in \cref{fig:rec_exp}.

The embeddings provided by \proposed{} consistently outperform the supervised baseline across various values of $K$ for both HR@$K$ and NDCG@$K$.
We computed the performance gain across all settings, observing an average improvement of 104\% when using embeddings from \proposed{}.
This outcome highlights the effectiveness of the pre-training phase, showing that the rich representations learned by \proposed{} successfully transfer to specialized downstream applications.

\subsection{Embedding Visualization}
\label{sec:visual}
\begin{figure}[htp]
\centerline{
\includegraphics[width=0.99\linewidth]{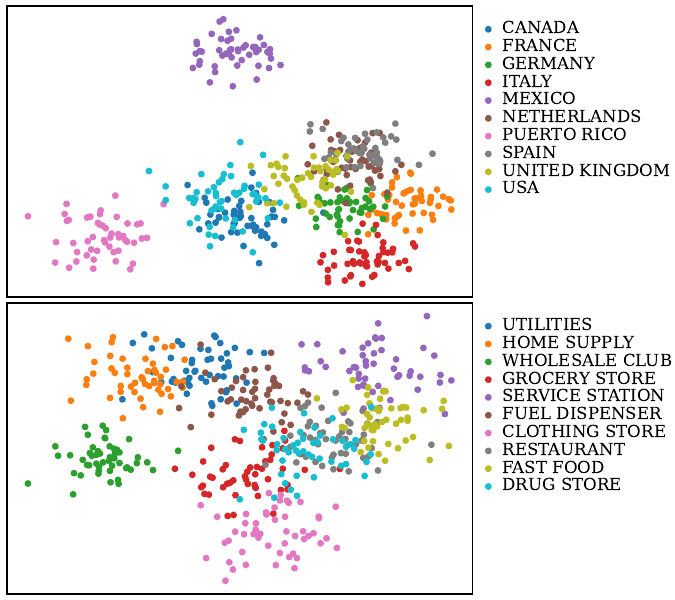}
}
\caption{
The \textit{merchant embedding} captures both \textit{location} and \textit{merchant category} information.
}
\label{fig:mrch_emb}
\end{figure}

In addition to objectively verifying the usefulness of the embeddings provided by \proposed{}, this subsection qualitatively explores the information they capture.
To this end, we visualize individual merchant embeddings by linearly projecting them from their high-dimensional space into a 2D plane.
We generated two distinct visualizations to analyze different relational aspects.
For the first plot, which focuses on geographic proximity, we created a sample of 500 merchants by randomly selecting 50 from each of 10 popular countries/regions.
For the second plot, focusing on business similarity, we created a sample of 500 merchants by selecting 50 from each of 10 popular merchant categories.
The resulting scatter plots are presented in \cref{fig:mrch_emb}; the top plot visualizes the location-based sample, while the bottom visualizes the category-based sample.

Observing the plot for the location-based sample (\cref{fig:mrch_emb}\textit{.top}), we see that merchants from the same countries or regions cluster together.
These individual clusters also form larger regional super-clusters.
More specifically, continental European countries form a large, distinct cluster.
The United States and Canada form another, with the United Kingdom located between this North American group and the continental European cluster.
Puerto Rico and Mexico form their own cluster, positioned near the one containing the United States and Canada.

Likewise, the plot for the category-based sample (\cref{fig:mrch_emb}\textit{.bottom}) reveals logical groupings based on business type.
For instance, drug stores and grocery stores are in close proximity, an intuitive finding that reflects the modern retail strategy where grocery stores incorporate pharmacies and drug stores expand their food and household staple selections.
Similarly, fast-food and full-service restaurants are clustered together, reflecting their shared role as food providers.
In contrast, wholesale clubs and clothing stores form distinct clusters that are more distant from the others, likely reflecting their specialized business models; for example, wholesale clubs offer a broad range of goods while clothing stores serve a narrow, specific market.
These visual findings suggest that the embeddings learned by \proposed{} capture meaningful geographic and categorical relationships between merchants, reflecting real-world market structures and consumer patterns.

\begin{figure}[htp]
\centerline{
\includegraphics[width=0.99\linewidth]{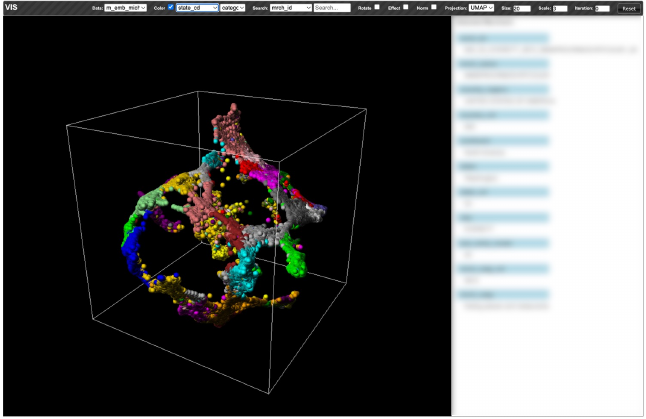}
}
\caption{
We developed a GUI to explore the embedding space. 
In the figure, merchant embeddings are visualized and color-coded according to their locations. 
Please note that the meta-information for each merchant shown in the right panel is blurred to protect privacy.
}
\label{fig:embedding_gui}
\end{figure}

In light of multiple existing works that leverage visualization and interaction for improved embedding analysis and interpretation~\cite{zheng2023embeddingtree, rathore2024verb}, we developed a graphical user interface (GUI) to visualize embeddings in 3D.
A screenshot of this tool is presented in \cref{fig:embedding_gui}.
The GUI allows for the interactive projection of embeddings into a 3D space using various algorithms (e.g., PCA~\cite{pearson1901liii}, $t$-SNE~\cite{van2008visualizing}, UMAP~\cite{mcinnes2018umap}) and supports dynamic coloring of points based on their associated metadata.
This interface has proven to be an effective tool for rapidly generating and testing hypotheses about the latent structures within the embedding space.

\subsection{Data and Model Scaling Analysis}
\label{sec:scaling_exp}
In this section, we analyze the impact of scaling on \proposed{}'s performance along two axes: training dataset size and model size.
To study data scaling, we trained models on datasets of varying sizes, created by sampling transactions from an increasing number of distinct cardholders.
For model scaling, we varied the model's size by adjusting its hidden dimensions.
Each resulting model's performance was evaluated using an index that quantifies its relative improvement over the currently deployed system on critical payment-related tasks.

\begin{figure}[htp]
\centerline{
\includegraphics[width=0.99\linewidth]{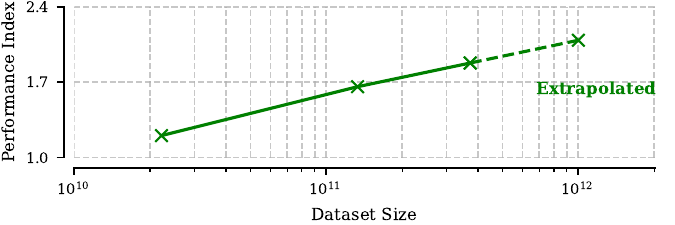}
}
\caption{
Model performance scales with dataset size, with an extrapolated point for a trillion-sized dataset.
}
\label{fig:scaling_data}
\end{figure}

First, we examine the impact of data scaling, with results depicted in \cref{fig:scaling_data}.
The figure clearly shows a positive correlation between the training dataset size and model performance.
This result strongly suggests that further increasing the dataset scale is a promising avenue for developing an even more powerful foundation model for transaction data.

\begin{figure}[htp]
\centerline{
\includegraphics[width=0.99\linewidth]{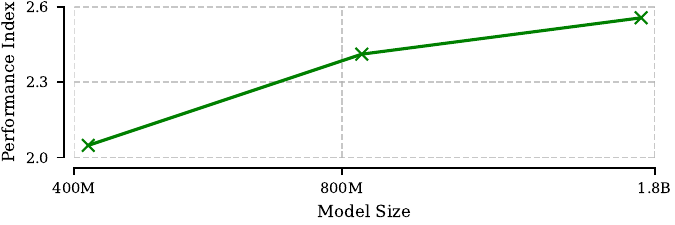}
}
\caption{
Performance scaling with model size, using 16-bit parameter precision.
}
\label{fig:scaling_model}
\end{figure}

Next, we analyze the effects of model scaling, as shown in \cref{fig:scaling_model}.
Increasing the model size also benefits the performance of \proposed{}.
However, unlike the trend observed with data scaling, the performance gains from increasing model size appear to diminish and eventually saturate.
This suggests that for a given dataset size, there may be a point of diminishing returns for simply increasing model parameters without a corresponding increase in data.

\section{Conclusion}
In this paper, we outlined the design of \proposed{}, a foundation model specifically engineered for transaction data.
\proposed{} simultaneously captures both consumer behavior and payment network signals (e.g., response codes, system flags), providing information crucial for applications such as accurate recommendation systems and abnormal behavior detection.
\proposed{} has demonstrated promising performance as a foundation model, both when utilized as a standalone system and as an embedding provider.
As a standalone abnormal behavior detection system, \proposed{} outperformed the currently deployed system by 111\%.
When leveraged as an embedding provider, its generated embeddings improved recommendation model performance by 104\%.
The insights derived from developing \proposed{} can inform the creation of foundation models for tabular data in other domains.
Future work includes several promising directions: enhancing \proposed{}'s performance through continued scaling~\cite{yeh2022embedding} and training strategy refinements; exploring graph-based modeling approaches to leverage multi-hop relationships between entities (e.g., cards interacting with shared merchants~\cite{yeh2022embedding}; investigating optimization techniques such as quantization and efficient attention mechanisms to reduce inference latency; integrating \proposed{} into LLM-based systems to enhance their capabilities in handling transaction data~\cite{fan2025enhancing,yeh2025empowering}; and adopting in-context learning to combat data drift~\cite{yeh2025tict}.

\bibliographystyle{ACM-Reference-Format}
\bibliography{section/reference}

\end{document}